\documentclass[runningheads]{llncs}
\usepackage{graphicx}
\usepackage[utf8]{inputenc}
\usepackage[english, russian]{babel}

\usepackage{graphicx}
\graphicspath{ {images/} }

\usepackage{caption} 
\usepackage{indentfirst}

\usepackage{amsmath}
\usepackage{amsbsy} 

\usepackage{multirow}

\begin{document}

\selectlanguage{english}

\title{DaNetQA: a yes/no Question Answering Dataset for the Russian Language}
\titlerunning{DaNetQA}
%
 \author{ 
Taisia Glushkova\thanks{The first two authors have equal contribution.} \inst{1}\orcidID{0000-0003-3521-3834} \and
Alexey Machnev \inst{1}\orcidID{0000-0001-5828-9810} \and
Alena Fenogenova\inst{2} \orcidID{0000-0003-3139-1668} \and
Tatiana Shavrina \inst{2} \orcidID{0000-0002-6976-0185} \and
Ekaterina Artemova \inst{1}\orcidID{0000-0003-4920-1623} \and \\
Dmitry I. Ignatov\inst{1}\orcidID{0000-0002-6584-8534}}
 \authorrunning{Glushkova et al.}
 \institute{National Research University Higher School of Economics, Moscow, Russia
 \url{https://cs.hse.ru/en/ai/computational-pragmatics/} 
\and
Sberbank, Moscow, Russia
\\
 \email{toglushkova@edu.hse.ru}  \\
 \email{\{echernyak,amachnev,dignatov\}@hse.ru}\\
 \email{\{Fenogenova.A.S,Shavrina.T.O\}@sberbank.ru}
 }

\maketitle              
\begin{abstract}

DaNetQA, a new  question-answering corpus, follows BoolQ~\cite{clark2019boolq}
 design: it comprises natural yes/no questions. Each question is paired with a paragraph from Wikipedia and an answer, derived from the paragraph. The task is to take both the question  and a paragraph as input and come up with a yes/no answer, i.e. to produce a binary output. In this paper, we present a reproducible approach to DaNetQA creation and investigate transfer learning methods for task and language transferring. For task transferring we leverage three similar sentence modelling tasks: 1) a corpus of paraphrases, Paraphraser, 2) an NLI task, for which we use the Russian part of XNLI, 3) another question answering task, SberQUAD. For language transferring we use English to Russian translation together with multilingual language fine-tuning.

\end{abstract}
\section{Introduction}
The creation of new datasets, aimed at new, challenging tasks, describing complex phenomena, related to various aspects of language understanding and usage, is core to the current view of modern language technologies. However, the majority of the datasets, created and published at the best venues, target the English-language tasks and cultural aspects, related to English-speaking society. 

In response to the bias towards English, new datasets and benchmarks are developed, that comprise multiple languages. One of the well-known examples of such multilingual datasets is XNLI \cite{conneau2018xnli}, which is a natural language inference dataset. Although this dataset is developed for 15 languages, including low-resource ones, the approach to its creation still excessively utilizes English data: the dataset entries are manually translated from English to other languages, without any specific adjustments. Although the translation-based approach is quick and dirty and allows us to overcome the lack of dataset for any language, other concerns arise. From a general point of view, we understand, that  translating is different from natural everyday language usage \cite{hickey1998pragmatics}. Thus translated datasets may have different statistics and word usage in comparison to text, composed from scratch. This may affect the quality of the models, trained further on the translated datasets, when applied to real-life data.

Another approach to dataset creation involves collecting datasets following the guidelines and annotation schemes, designed for original datasets in English. RuRED \cite{rured} is a recent example of such a dataset: it is created following TACRED \cite{zhang2017tacred} annotation scheme and collection pipeline. This approach however is criticized for the obvious lack of novelty. 

Nevertheless, in this paper we stick to the second approach to the dataset creation and more or less follow the pipeline, developed for BoolQ \cite{clark2019boolq}, to create a new dataset for binary questions in Russian, which we refer to as DaNetQA \footnote{The dataset is available at: \url{https://github.com/PragmaticsLab/DaNetQA}}. We only deviate from the BoolQ pipeline, if we do not have access to proprietary data sources and instead use  crowdsourcing.  The motivation to re-create BoolQ lies, first, in the lack of question-answering dataset for Russian, and second, in the fact, that binary question answering appears to be a more challenging task, when compared to SQuAD-like  and natural language inference tasks \cite{wang2019superglue}. Thus we hope that DaNetQA may become of great use both for chat-bot technologies, which massively use question answering data, and for a thorough evaluation of deep contextual encoders, such as BERT \cite{devlin2019bert} or XLM-R \cite{conneau2019unsupervised}. 

As the annotations for DaNetQA are crowd-sourced and require payment, we explore different strategies that can help to increase the quality without the need to annotate larger amounts of data. This leads us to two strategies of transfer learning: transferring from tasks, which have a similar setting, and transferring from English, keeping in mind, that DaNetQA recreates BoolQ, but in a different language. 
Our main contributions are the following:
\begin{enumerate}
    \item We create and intent to publish in open access a new middle-scale dataset for the Russian language, DaNetQA, which comprises yes/no questions, paired with paragraphs, providing enough information to answer the questions (Subsection~\ref{subsec:collection}) and report its statistics (Subsection~\ref{subsec:stats});
    \item We establish a simple baseline and a more challenging deep baseline for DaNetQA (Subsection~\ref{subsec:stats});
    \item We explore the applicability of  transfer learning techniques, of which some overcome the established baseline (Section~\ref{sec:experiments}). 
\end{enumerate}


The paper is organised as follows. Section~\ref{sec:dataset} describes the data collection process and provides the reader with their basic statistics. In Section~\ref{sec:experiments}, all the conducted experiments are described. Section~\ref{sec:res} discusses the obtained results. In Section~\ref{sec:relwork}, related work is summarised. Section~\ref{sec:concl} concludes the paper.

\section{Dataset} \label{sec:dataset}

\subsection{Collection} \label{subsec:collection}
Our approach to DaNetQA dataset creation is inspired by the work of \cite{clark2019boolq}, where the pipeline from NQ \cite{kwiatkowski2019natural} is used as a base. 

Questions are generated on the crowdsourcing platform Yandex.Toloka, which is a good source for Slavic languages data generation compared to other language groups due to the origin of the platform. Questions are created by crowd workers following an instruction, that suggests phrases that can be used to start a phrase – a broad list of templates (more than 50 examples). 

Generated yes/no questions are then treated as queries in order to retrieve relevant Wikipedia pages with the use of Google API. Questions are only kept if 3 Wikipedia pages could be returned, in which case the question and the text of an article are passed forward to querying a pre-trained BERT-based model for SQuAD to extract relevant paragraphs.

Finally, crowd workers label question and paragraph pairs with ``yes'' or ``no'' answers, and the questions that could not be answered based on the information form the paragraph are marked as ``not answerable''. Annotating data in such a manner is quite expensive since not only crowd workers read through and label thousands of pairs manually, but there is also used a high overlap of votes to ensure the high quality of the dataset. In addition, Google API also requires some payments to process high numbers of queries.

The final labels for each pair are picked based on the majority of votes. In case of uncertainty when there is no label picked by the majority, the pair is being checked and labeled manually by the authors. 

Aside from the overlap used on the final step, we use a number of gold-standard control questions that are randomly mixed into the tasks to make sure that crowd workers do the annotation responsibly. Also, each annotator goes through a small set of learning tasks before starting labelling the actual pairs.


\subsection{Statistics} \label{subsec:stats}

In the following section, we analyze our corpus to better understand the nature of the collected questions and paragraphs and compute all kinds of descriptive statistics. Statistics presented below, include the minimum, average, and maximum length of questions and paragraphs in tokens (Table \ref{table1}) and the distribution of text lengths in the entire dataset as a whole. The total number of yes/no questions in Dev/Test/Train sets can be found in Table \ref{table2}.

\begin{table}[!ht]
	\captionsetup{justification=centering}
	\caption{Descriptive statistics of collected question/paragraph pairs (in tokens).}
	\label{table1}
	\centering
	\begin{tabular}{ p{0.19\textwidth} | p{0.09\textwidth}| p{0.09\textwidth}| p{0.09\textwidth} | p{0.09\textwidth} | p{0.09\textwidth} | p{0.09\textwidth} | p{0.09\textwidth}}
    	\hline
    	 & \hfil Count & \hfil Mean &  \hfil Min & \hfil 25\% & \hfil 50\% & \hfil 75\% &  \hfil Max \\
    	\hline 
    	Questions & \hfil2691 & \hfil5 & \hfil2 & \hfil5 & \hfil5 & \hfil6 & \hfil14 \\ 
    	Paragraphs & \hfil2691 & \hfil90 & \hfil37 & \hfil72 & \hfil88 & \hfil106 & \hfil206 \\ 
    	\hline
	\end{tabular}
\end{table}

\begin{table}[!ht]
	\captionsetup{justification=centering}
	\caption{The number of yes/no questions in Dev, Test and Train sets. Symbol \# stands for the total number of yes (or no) questions (columns 3 and 5) and \% stands for the ratio of yes (or no) questions in the dataset (columns 4 and 6).}
	\label{table2}
    \centering
	\begin{tabular}{ p{0.13\textwidth} | p{0.13\textwidth}| p{0.13\textwidth}| p{0.13\textwidth} | p{0.13\textwidth} | p{0.13\textwidth} }
		\hline
		 & \hfil Size & \hfil \#Yes & \hfil Yes\% & \hfil \#No & \hfil No\%  \\
		\hline
		Dev set & \hfil821 & \hfil672 & \hfil81.8 & \hfil149 & \hfil18.1  \\ 
		Test set & \hfil805 & \hfil620 & \hfil77 & \hfil185 & \hfil22.9  \\ 
		Train set & \hfil1065 & \hfil805 & \hfil75.5 & \hfil260 & \hfil 24.4  \\
		\hline
	\end{tabular}
\end{table}	

\begin{figure}[!ht]
	\centering
	\includegraphics[width=11cm]{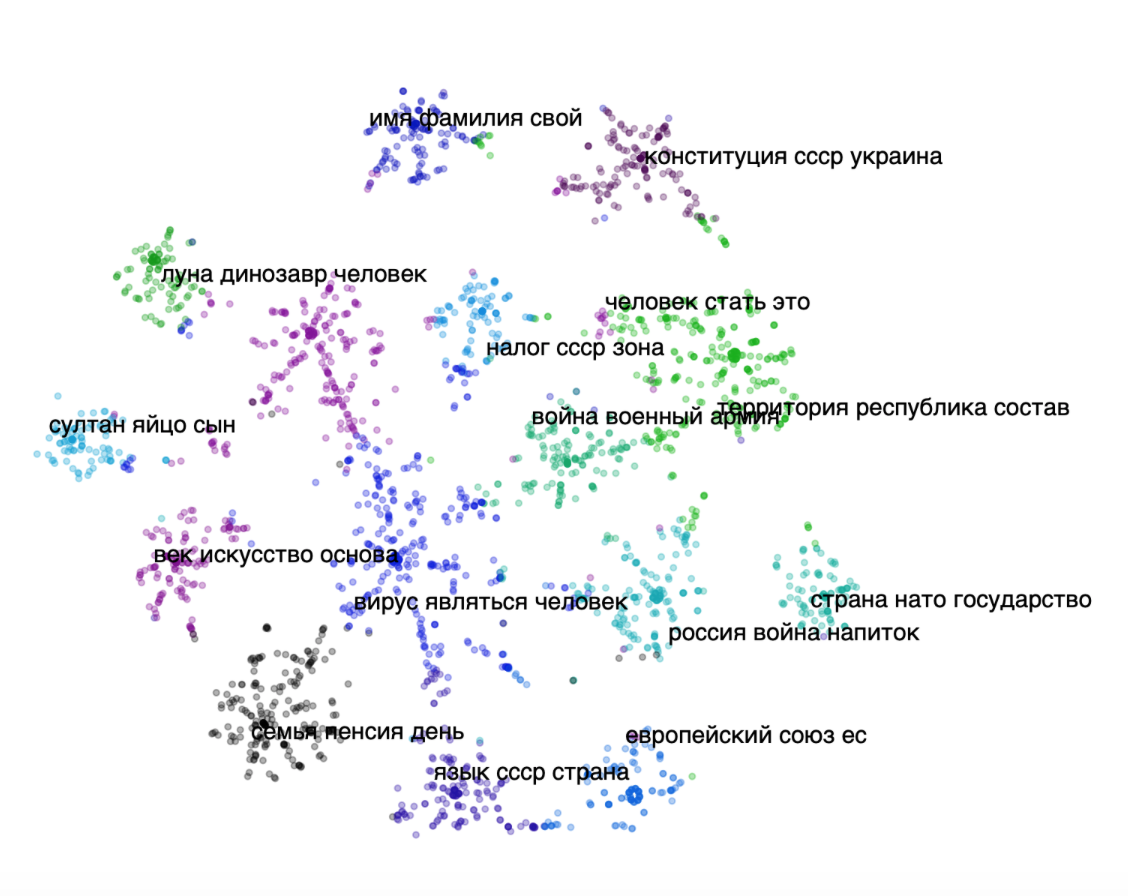}
	\caption{The t-SNE visualization of an LDA model for 15 topics, trained on concatenated questions and paragraphs.}
	\label{img1}
\end{figure}

We also provide a brief overview of categories covered by our question and paragraph pairs (Table \ref{table3}) and the t-SNE visualization of an LDA model for 15 topics, trained on concatenated questions and paragraphs (Figure \ref{img1}). Each topic is labelled with top-3 tokens that represent a topic summary.

\begin{table}[!ht]
	\captionsetup{justification=centering}
	\caption{Question categorization of DaNetQA. Top-10 question topics.}
	\label{table3}
	\selectlanguage{russian}
    \begin{tabular}{ p{0.3\textwidth} |  p{0.45\textwidth} | p{0.1\textwidth} | p{0.07\textwidth} }
		\hline
		\hfil Category & \hfil Example & \hfil Percent & \hfil Yes\%  \\
		\hline 
		Война (War/Military) & Был ли взят калинин немцами? & \hfil 10.51 & \hfil 54.7 \\
		История (History) & Были ли фамилии у крепостных крестьян? & \hfil 9.55 & \hfil 87.1 \\
		Космос (Space/Galaxy) & Есть ли жизнь на других планетах солнечной системы? & \hfil 9.10 & \hfil 79.5 \\
		СССР (USSR) & Была ли конституция в СССР? & \hfil 7.61 & \hfil 83.9 \\ 
		Здоровье (Health) & Передаётся ли чумка от кошки к человеку? & \hfil 7.17 & \hfil 84.9 \\ 
		Семья (Family) & Был ли у гагарина брат? & \hfil 6.76 & \hfil 81.3 \\ 
		Искусство (Art/Literature) & Существует ли жанр эклектика в живописи? & \hfil 6.54 & \hfil 75.5 \\
		Евросоюз (EU) & Входит ли чехия в евросоюз? & \hfil 6.28 & \hfil 86.9 \\
		География (Geography) & Была ли албания в составе югославии? & \hfil 6.05 & \hfil 76.6 \\ 
		Связи (Communications) & Был ли лермонтов знаком с пушкиным? & \hfil 5.42 & \hfil 82.8 \\
		\hline
	\end{tabular}
\end{table}	

Since the crowd workers were provided (but not restricted) with question starter templates, in order to generate yes/no questions, we decided to check the frequency of the used bigrams that the generated questions begin with (see Table \ref{table4}).

\begin{table}[!ht]
	\captionsetup{justification=centering}
	\caption{Top-15 frequent bigrams, that start the questions.}
	\label{table4}
	\centering
	\selectlanguage{russian}
	\begin{tabular}{ p{0.7\textwidth} |  p{0.15\textwidth} }
		\hline
		\hfil Indicator words & \hfil Frequency\\
		\hline 
		был ли (was (\textit{3rd pers., sing., masc.}) there) & \hfil 575 \\
		есть ли (is (\textit{3rd pers., sing., masc.}) there) & \hfil 362 \\ 
		была ли (was (\textit{3rd pers., sing., fem.}) there) & \hfil 317 \\ 
		были ли (was (\textit{3rd pers., plur}) there) & \hfil 302 \\ 
		входит ли (is included (\textit{3rd pers., sing.})) & \hfil 201  \\ 
		едят ли (do [they] eat (\textit{3rd pers., plur})) & \hfil 130 \\ 
		правда ли (is [it] true) & \hfil 130 \\ 
		разрешено ли (is [it] allowed (\textit{3rd pers., sing., neutr.})) & \hfil 74 \\ 
		передаётся ли (is [it] transmitted (\textit{3rd pers., sing.})) & \hfil 70 \\ 
		состоит ли (does [it] consist (\textit{3rd pers., sing.}) in) & \hfil 68 \\
		бывает ли (is (\textit{3rd pers., sing., fem.}) there) & \hfil 55 \\ 
		вреден ли (is [it] harmful (\textit{3rd pers., sing., masc.}))& \hfil 55 \\ 
		вредна ли (is [it] harmful (\textit{3rd pers., sing., fem.})) & \hfil 50 \\ 
		существует ли (does [it] exist (\textit{3rd pers., sing.}))& \hfil 41 \\ 
		а была (was (\textit{3rd pers., sing., fem.}) there) & \hfil 30 \\
		\hline
	\end{tabular}
\end{table}	  

Note that this list of starter phrases is consistent with the one in the \cite{clark2019boolq} pipeline. Indeed, in the paper, authors select questions that begin with a certain set of words (``did'', ``do'', ``does'', ``is'', ``are'', ``was'', ``were'', ``have'' , ``has'', ``can'', ``could'', ``will'', ``would''). Our list and the list used in the creation of BoolQ are somewhat similar, if we omit the particle \foreignlanguage{russian}{``ли''}, which is added in Russian when creating interrogative sentences. The majority of the questions in the corpus are starting with these words.
\selectlanguage{english}

\section{Experiments} \label{sec:experiments}

The DaNetQA task is formulated as a binary classification task. The input to the model consists of the question and paragraph pair. The task is to return either 0 or 1, such that the positive answer indicates that the answer to the question, based on the information from the paragraph is ``yes''. Otherwise, 0 stands for the negative answer. We evaluated two baseline algorithms, to which multiple transferring techniques are compared. As the source for the transfer, we used either other tasks, which have a similar setting, or sources in other languages.

\subsection{Baseline} \label{subsec:baseline}


We exploit two baseline approaches:

\begin{enumerate}
    \item FastText for binary classification \cite{joulin2017bag}; 
    \item A deep learning approach: fine-tuning RuBERT model on  DaNetQA.
\end{enumerate}

\textbf{FastText.} The FastText classifier trained on the concatenated vectors of questions and paragraphs with the use of pre-trained vector representations (taken from the official website) showed slightly lower accuracy, than the one trained only on the DaNetQA dataset (Table~\ref{table:table_baseline}).

\textbf{BERT.} Fine-tuning DeepPavlov RuBERT \cite{kuratov2019adaptation} on the imbalanced DaNetQA, leads to resulting model being uncalibrated. Therefore the predicted probability for the positive class is higher in most cases. To tackle this problem, we aim to find an optimal predicted probability threshold for validation dataset. This is done in two steps:

\begin{enumerate}
    \item For each threshold $t$ in some discrete subset of range $[0, 1]$, ``yes" answer $precision(t)$ and $recall(t)$ calculated;
    \item $F_1(t) = \dfrac{2 \cdot precision(t) \cdot recall(t)}{precision(t) + recall(t)}$ is calculated;
    \item Threshold $t$ for the largest $F_1(t)$ is selected.
\end{enumerate}

We train a model for 5 epochs, saving checkpoint each 40 training steps, and then select the best checkpoint according to the accuracy value. For this task, we set the learning rate to 3e-5, and linearly decrease it up to zero to the end of the training.

\begin{table}[!ht]
	\captionsetup{justification=centering}
	\caption{Accuracy for binary classification with FastText.}
	\label{table:table_baseline}
	\centering
	\begin{tabular}{ p{0.1\textwidth} |  p{0.15\textwidth} | p{0.25\textwidth} |  p{0.15\textwidth}   }
		\hline
		 &  \hfil FastText & \hfil pre-trained FastText & \hfil RuBERT \\
		\hline 
		ACC & \hfil 0.81366 & \hfil 0.80745 & \hfil 0.7975  \\ 
		 \hline
	\end{tabular}
\end{table}

FastText can be seen as a strong baseline, as it manages to achieve results comparable to the larger RuBERT model.  Further, we will refer to the RuBERT baseline to compare with transfer learning techniques.

\subsection{Task Transferring}





To transfer learning from other tasks, we make two steps: 
\begin{enumerate}
    \item Fine-tuning the pre-trained transformer model on a similar task;
    \item Fine-tuning the model on DaNetQA.
\end{enumerate}
The tasks, used for the first step, are Paraphraser, the Russian part of XNLI, and Task A of the SberQUAD. For each task, we initialize model weights with DeepPavlov RuBERT. All three datasets are used independently.


\textbf{Paraphraser} \cite{pronoza2015construction} is a dataset for the paraphrasing task: it consists of sentence pairs, each of which is labeled as paraphrase, not paraphrase or maybe paraphrase. This task is close to DaNetQA as the model is required to detect linkage between sentences. We transform this task into a binary classification problem, where the label is true only for definitely paraphrase. 

 \textbf{XNLI} \cite{conneau2018xnli} is the dataset for the language inference task, translated to 15 languages, including Russian. The task is to predict if one sentence is entailment or contradiction of another, or two sentences are neutral. Each pair has initial annotation and five other crowd workers annotations. For our task, we kept only the Russian part of the dataset and only sentences where the majority label is the same as initial. As for Paraphraser, we transformed the task into a binary classification problem, and the only entailment is treated as the positive class.

 \textbf{SberQUAD task A} is the part of the Sberbank question-answer challenge. In task A, a model needs to predict if a given text contains an answer to a given question.

Since each task in the first step is, as DaNetQA, a binary classification problem, we use the same approach with training and selecting the best model, as for fine-tuning BERT on DaNetQA in the baseline approach. The best model selection and evaluation of further DaNetQA fine-tuning is also the same as in the baseline solution.

The model is fine-tuned on the first-step task with initial learning rate 3e-5 and then fine-tuned on DaNetQA with the same initial learning rate, linearly decreasing to zero in each case. During training, a checkpoint is saved each 40 steps for Paraphraser, 75 steps for XNLI, and 3000 steps for SberQUAD, then the best one is selected by accuracy value on the pre-training dataset. We perform 5 different pre-training runs on each dataset, then fine-tune each pre-trained model 5 different times on DaNetQA. We show mean and standard deviation values over different fine-tuning runs after the best pre-training run, selected by mean accuracy value.

\subsection{Language Transferring}





Similarly to  task transferring, for language transferring  we fine-tune the transformer on a similar dataset and then further fine-tune it on DaNetQA. The difference is that the dataset in the first-step task is not in Russian, so we should train the model on its machine translation or use a multilingual pre-trained model. We tested the following configurations:

\begin{itemize}
    \item Fine-tune RuBERT on translated BoolQ and test on DaNetQA;
    \item Fine-tune RuBERT on translated BoolQ and then fine-tune and test on DaNetQA;
    \item Fine-tune XLM-R on BoolQ and then fine-tune on DaNetAQ.
\end{itemize}

The hyperparameters and algorithm of the best model selection and evaluation are the same as in the baseline approach for fine-tuning RuBERT on DaNetQA, apart from the initial learning rate for pre-training and fine-tuning XLM-R model is 1e-5.


Intuitively, when pre-trained on BoolQ, which inspired DaNetQA, the model should have the relevant knowledge to solve our task. As the BoolQ dataset is in English, we should use some kind of multilingual technique to enable transfer learning. The core methods in this case are:
\begin{enumerate}
    \item To use machine translation models to translate either BoolQ to Russian, or DaNetQA to English, and fine-tune a monolingual model on BoolQ and further use it for DaNetQA;
    \item To use a multi-lingual model and BoolQ for pre-training in a straightforward way.
\end{enumerate}

To translate BoolQ into Russian we used the Helsinki-NLP/opus-mt-en-ru machine translation model from HuggingFace Transformers\footnote{\url{https://huggingface.co/transformers/}} library.

\textbf{XLM-R} \cite{conneau2019unsupervised} is a transformer-based model, trained on different languages. This model should capture semantic information in higher layers, so fine-tuning on English data could be good initialization to process further DaNetQA.

\section{Results} \label{sec:res}

To evaluate the transfer learning approaches and compare against baselines, we calculated accuracy and $F1$ scores for ``Yes'' answers. However, since ``Yes'' is the most common answer, we should pay more attention to ``No'' answers, which are less frequent. Therefore we report precision-recall AUC for ``No'' answers. To understand how sensitive is the model we calculated ROC-AUC. The result values of each configuration are described in Table \ref{table:ResultsTable}.

\begin{table}[htp!]
	\captionsetup{justification=centering}
	\caption{Results of the task solving approaches.}
	\label{table:ResultsTable}
	\centering
\begin{tabular}{ c c c c c c }
\hline
 \multicolumn{1}{p{3cm}}{\centering Transfer dataset} & 
 \multicolumn{1}{p{1cm}}{\centering Model} & 
 \multicolumn{1}{p{1.4cm}}{\centering Fine-tune on DaNetQA} & 
 \multicolumn{1}{p{2cm}}{\centering ACC} & 
 \multicolumn{1}{p{2cm}}{\centering ROC - AUC} & 
 \multicolumn{1}{p{2cm}}{\centering ``No" precision  -recall AUC}  \\
\\
\hline

- & RuBERT & Yes   &  $82.46 \pm 0.48$  &  $75.92 \pm 0.38$  &  $57.16 \pm 2.38$ \\
- & XLM-R & Yes   &  $81.44 \pm 1.28$  &  $67.76 \pm 2.53$  &  $49.41 \pm 3.43$ \\
\hline
Paraphraser & RuBERT & Yes   &  $82.48 \pm 0.53$  &  $75.02 \pm 0.48$  &  $55.22 \pm 1.97$ \\
XNLI (ru) & RuBERT & Yes    &  $81.89 \pm 0.76$  &  $78.56 \pm 1.99$  &  $57.15 \pm 2.47$ \\
SberQUAD Task A & RuBERT & Yes   &  $82.63 \pm 0.88$  &  $\pmb{80.74 \pm 1.16}$  &  $\pmb{63.02 \pm 1.54}$ \\
\hline
BoolQ translated & RuBERT & No   &  $74.04 \pm 1.88$  &  $62.74 \pm 2.61$  &  $32.64 \pm 2.86$ \\
& RuBERT & Yes   &  $82.56 \pm 0.29$  &  $\pmb{80.81 \pm 1.50}$  &  $62.49 \pm 1.84$ \\
BoolQ & XLM-R & No   &  $78.68 \pm 0.97$  &  $66.53 \pm 3.21$  &  $42.25 \pm 4.65$ \\
 & XLM-R & Yes   &  $\pmb{83.20 \pm 0.30}$  &  $79.55 \pm 1.63$  &  $61.84 \pm 1.00$ \\

\end{tabular}
\end{table}


Three Russian datasets in the task transferring approach help the model to achieve higher quality than without fine-tuning. Nevertheless, the highest improvement is achieved when transferring from SberQUAD Task A in line with similar experiments of \cite{clark2019boolq}.  As the tasks differ from Yes/No question answering, results are still lower than those, achieved by the language transferring approach.

Our results show, that  fine-tuning the XLM-R model on BoolQ (both translated and not) and using the model for further evaluation on DaNetQA, fails in comparison to other techniques. A possible explanation for this is the difference in the topics, covered by both datasets. However, if the model is further fine-tuned on DaNetQA, the superior results are achieved due to the cross-lingual nature of the model. 

\section{Related work} \label{sec:relwork}
\textbf{Binary question answering} is  a significant part of machine reading comprehension problem. Binary questions are present in the following datasets in English : CoQA \cite{reddy2019coqa}, QuAC \cite{choi2018quac}, HotPotQA \cite{yang2018hotpotqa} and ShARC \cite{saeidi2018interpretation}. Some of these datasets, in particular, HotPotQA require multi-hop reasoning, for which answering binary questions is crucial. However BoolQ \cite{clark2019boolq} to the best of our knowledge is the only dataset, devoted to binary questions exclusively. 

\textbf{Transfer learning} is one of the leading paradigms in natural language processing. It leverages pre-training on large-scale datasets as a preliminary step before fine-tuning a contextualized encoder for the task under consideration. Applications of transfer learning range from part-of-speech tagging \cite{kim2017cross} to machine translation \cite{ji2020cross}. As the language modelling can be seen as independent tasks, some researchers see pre-training with language modelling objective as a part of the transfer learning paradigm \cite{howard2018universal,golovanov2019large}. Pre-training on natural language understanding tasks, in particular, on sentence modelling tasks, help not only to improve the quality of the task under consideration \cite{clark2019boolq,shang2019find,kamath2019pre}, but also to derive semantically meaningful sentence embeddings that can be compared using cosine-similarity \cite{reimers2019sentence}.

\textbf{New datasets for the Russian language} are rarely published. Although Russian is one of the most widely spoken languages, the amount of datasets, suitable for NLP studies is quite limited. Dialogue, AIST and AINL conferences are the main venues for the Russian datasets to appear. Among recent datasets are RuRED \cite{rured} and SentiRusColl \cite{kotelnikova2019sentiruscoll}. One more example of the previous studies with a large collection of question queries in Russian and the associated results can be found in ~\cite{Volske:2015}.


%
%
\section{Conclusion} \label{sec:concl}
In this paper, a new question answering dataset, DaNetQA, is presented. It comprises binary yes/no questions, paired with paragraphs, which should be used to answer the questions. The overall collection procedure follows the design of the BoolQ dataset, which is a magnitude larger in size than DaNetQA, partially due to the use of proprietary sources. 
We establish a straightforward baseline, exploiting FastText and RuBERT models and experiment with multiple transfer learning settings. Our results show, that on the one hand, the English dataset can be leveraged to improve the results for the Russian one. However, we can not confirm, that we can re-use BoolQ for training the model while keeping DaNetQA for evaluation only. 
This brings us to the following conclusion: although the re-creation of English datasets in other languages may seem like a redundant and secondary activity, the current state of the cross-lingual models does not allow for perfect language transfer. It is not enough to train the model on the English data. It seems impossible to gain high-quality results if the model is not trained in the target language. This highlights the need for future development: the development of more advanced cross-lingual contextualized encoders as well as more sophisticated datasets to evaluate cross-lingual tasks. As for DaNetQA development, we plan to enlarge the dataset with more question-paragraph pairs and to extend the dataset with an unanswerable question, affecting though the task setting. 

\subsubsection*{Acknowledgements.}
This paper was prepared in the Laboratory for Model and Methods of Computational Pragmatics within the framework of the HSE University Basic Research Program and funded by the Russian Academic Excellence Project '5-100'. 

\bibliographystyle{splncs04}
\bibliography{papers}

\end{document}